\documentclass[10pt,twocolumn,letterpaper]{article} 
\usepackage{avss}
\usepackage{times}
\usepackage{epsfig}
\usepackage{graphicx}
\usepackage{amsmath}
\usepackage{amssymb}

\avssfinalcopy 
\def\avssPaperID{****} 

\ifavssfinal\fi
\begin{document}
\title{RepSFNet : A Single Fusion Network with Structural Reparameterization for Crowd Counting}

\author{
Mas Nurul Achmadiah\\
Department of Electro-Optical Engineering\\
National Formosa University\\
Yunlin, Taiwan\\
\tt\small {masnurul@polinema.ac.id}
\and
Chi-Chia Sun*\\
Department of Electrical Engineering\\
National Taipei University\\
New Taipei City, Taiwan\\
\tt\small {chichiasun@gm.ntpu.edu.tw}
\and
Wen-Kai Kuo\\
Department of Electro-Optical Engineering\\
National Formosa University\\
Yunlin, Taiwan\\
\tt\small {wkkuo@nfu.edu.tw}
\and
Jun-Wei Hsieh\\
College of Artificial Intelligence and Green Energy\\
National Yang Ming Chiao Tung University\\
Hsinchu City, Taiwan\\
\tt\small {jwhsieh@nctu.edu.tw}
}
 
\vspace{-2 mm}

\maketitle

\begin{abstract}
Crowd counting remains challenging in variable-density scenes due to scale variations, occlusions, and the high computational cost of existing models. To address this, we propose RepSFNet (Reparameterized Single Fusion Network), a lightweight architecture designed for accurate and real-time crowd estimation. RepSFNet combines large-kernel convolutional power with a efficient, suitable for low-power edge computing. The architecture includes three components: (i) a RepLK-ViT backbone using large reparameterized kernels for efficient multi-scale feature extraction; (ii) a Feature Fusion module that integrates ASPP and CAN for robust, density adaptive context modeling; and (iii) a Concatenate Fusion module to preserve spatial resolution and produce high-quality density maps. By avoiding attention mechanisms and multi-branch designs, RepSFNet reduces both parameters and FLOPs, enhancing runtime efficiency. The loss function combines Mean Squared Error (MSE) and Optimal Transport (OT), further improving count accuracy. Experiments on ShanghaiTech, NWPU, and UCF-QNRF show that RepSFNet delivers competitive accuracy with up to 34\% lower inference latency compared to P2PNet, M-SFANet, M-SegNet, STEERER, and Gramformer, making it more efficient and suitable for low-power edge computing.

\end{abstract}
\vspace{-5mm}
\section{Introduction}
Crowd counting is challenging due to extreme density variations, occlusions, and environmental changes like lighting and perspective distortion. These conditions require models that balance accuracy and computational efficiency for real-time deployment. While CNNs have improved performance significantly \cite{3}, many existing methods rely on attention or multi-branch designs that increase memory use and inference time. We propose RepSFNet (Reparameterized Single Fusion Network), a lightweight architecture designed for efficient and accurate crowd estimation. Built on a RepLK-ViT backbone with large reparameterized kernels, RepSFNet extracts rich multi-scale features with low overhead. It includes a stem block, four RepLK stages, and a Feature Fusion module that integrates ASPP \cite{assp} and CAN \cite{NR3} for adaptive context modeling. A Concatenate Fusion module maintains semantic and spatial consistency, producing high-resolution density maps. The network is trained with a loss function MSE + Optimal Transport loss to improve count accuracy. 
\vspace{1mm}
\newline
Experiments on ShanghaiTech, NWPU, and UCF-QNRF show that RepSFNet delivers competitive accuracy. Ablation studies also reveal up to 34\% lower latency compared to widely used baselines including M-SFANet, P2PNet, M-SegNet, and STEERER. Our contributions include (1) A RepLK-ViT-based backbone with reparameterized large kernels for efficient feature extraction. (2) A density-adaptive fusion module improving ASPP and CAN for context awareness. (3) A lightweight spatial-semantic fusion design that ensures detail and consistency in high-res outputs. The outline of the paper covers related work (Section \ref{related work}), method (Section \ref{method}), experiments (Section \ref{Experimental}), and conclusion.(Section \ref{Conclusions}).

\section{Related Works}
\label{related work}
\subsection{Large Kernel Design in CNNs}
Large kernel design in Convolutional Neural Networks (CNNs) has gained increasing attention due to its ability to effectively capture long-range dependencies and global context—critical for dense prediction tasks such as crowd counting. Traditional CNNs, such as those built on VGG-Net, primarily relied on small 3 $\times$ 3 kernels, limiting the receptive field and requiring deeper layers to model large contexts. To address this, various approaches have been explored, including stacked convolutions and dilated convolutions, but these often lead to increased computational costs and inefficiency.

\begin{figure}[!h]
    \centering
    \includegraphics[width=0.32\textwidth]{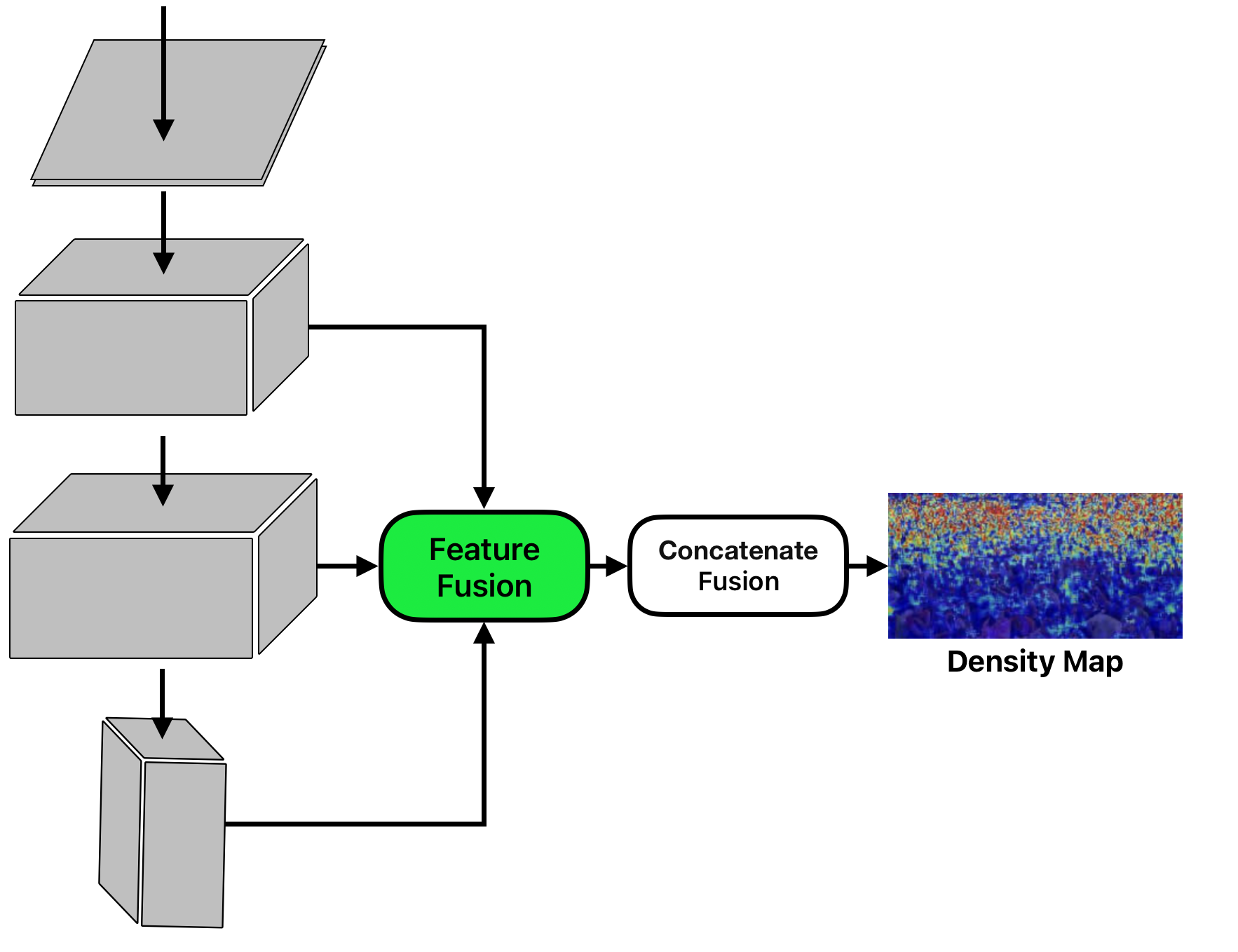}
    \caption{A Simple Single Fusion Network Architecture}
    \label{fig:SA}
\end{figure}

Inspired by the global receptive field of Vision Transformers (ViTs), recent works like RepLKNet \cite{NR1} propose reparameterized large kernel convolutions, offering a more efficient and scalable alternative to transformers. Additionally, convolutional methods such as CK-Conv and GFNet \cite{31} demonstrate that explicitly designing large kernel filters can enhance the modeling of spatially distributed patterns without relying solely on attention mechanisms. 
\begin{figure}[!h]
    \centering
    \includegraphics[width=8cm, height=7cm]{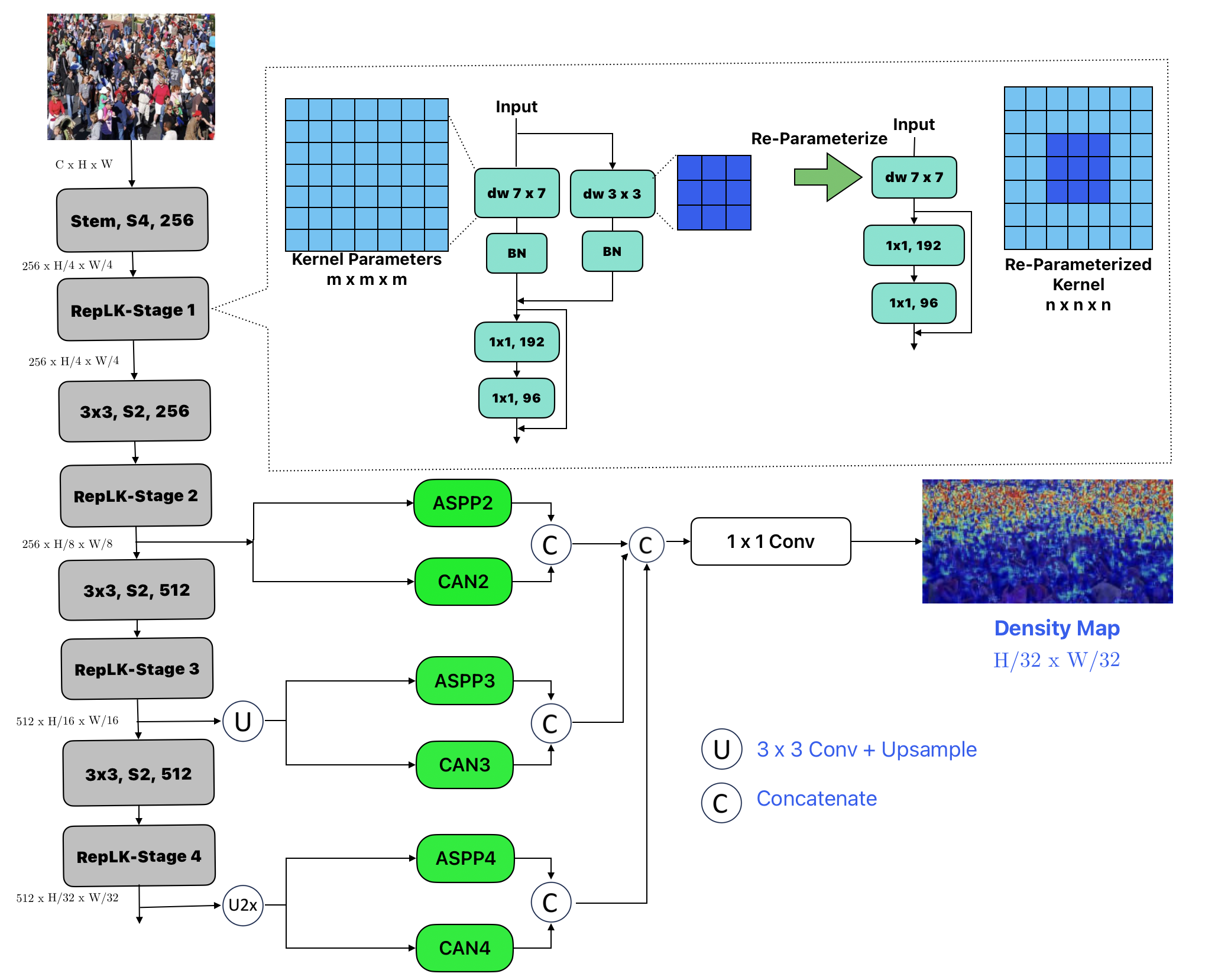}
   \caption{The complete architecture of RepSFNet}
    \label{fig:RepSFnet}
\end{figure}
\vspace{-2.5mm}

\subsection{Density Map Ground Truth}
In crowd counting, density maps represent the spatial distribution of people within an image. The density map ground truth is derived from point annotations marking individual locations and is used to train convolutional neural networks (CNNs) for estimating crowd density. Instead of directly predicting the total number of people, CNN-based models generate a density map where the integral over any region estimates the number of individuals in that area \cite{35}.
\vspace{-1mm}
\section{Method}
\label{method}
This subsection outlines our methodological framework. We start with the RepLK-ViT backbone, which offers efficient and scalable multi-scale feature extraction for crowd counting. Then, we present the Feature Fusion module for hierarchical context aggregation and enhanced spatial-semantic representation, followed by the Concatenate Fusion mechanism that preserves semantic consistency and spatial detail in density map generation. These components form RepSFNet—a unified and efficient architecture for crowd estimation. Figure \ref{fig:SA} illustrates the overall pipeline, highlighting the fusion modules, while Figure \ref{fig:RepSFnet} details the network structure and reparameterization process.
\vspace{-2mm}
\subsection{Reparameterizing Large Kernels}
The backbone architecture is shown in Figure \ref{fig:RepSFnet}, highlighting the RepLK-ViT structure, which uses the Reparameterized Large Kernel Vision Transformer (RepLK-ViT), a convolutional design for dense prediction tasks like crowd counting. It begins with a 4×4 convolutional stem (stride 4), followed by four RepLK stages with large kernels reparameterized into efficient 3×3 convolutions with batch normalization and pointwise layers. Each stage increases channels (256→512) and reduces resolution (H/4→H/32), enabling effective multi-scale feature extraction. Final features are refined using ASPP \cite{assp} and CAN \cite{NR3}, yielding a high-resolution output at H/32 × W/32. Although RepLK-ViT omits self-attention, its hierarchical design and large receptive fields reflect ViT principles. This design delivers transformer-like global perception with the efficiency of CNNs, reflecting a hybrid philosophy that balances contextual understanding and computational practicality.

\subsection{Feature Fusion and Concatenate Fusion}
\begin{figure}[!h]
    \centering
    \includegraphics[width=5.5cm, height=4.5cm]{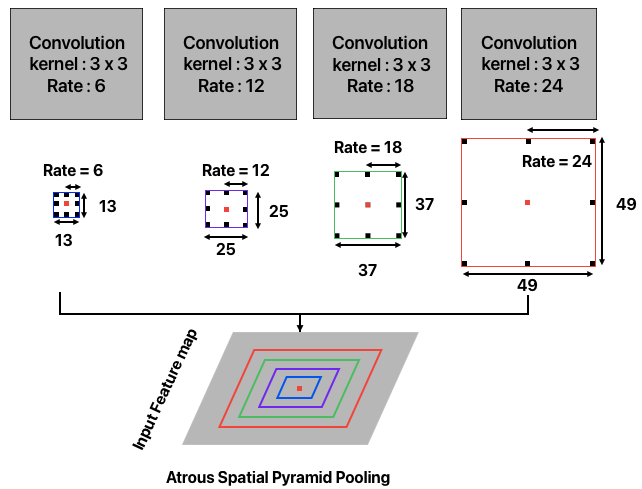}
    \caption{Atrous Spatial Pyramid Pooling}
    \label{fig:ASPP}
\end{figure}

The Feature Fusion module integrates Atrous Spatial Pyramid Pooling (ASPP) and Context-Aware Network (CAN) to improve multi-scale context modeling for crowd density estimation. ASPP uses parallel dilated convolutions (rates 6, 12, 18, 24), a 1×1 convolution, and global pooling to extract rich features across varied receptive fields, addressing scale variation (Figure \ref{fig:ASPP}). CAN modules refine spatial features by adaptively emphasizing relevant scales per pixel. They take inputs from RepLK stages and ASPP outputs to perform selective spatial weighting based on contrast. ASPP provides fixed-scale context, while CAN adds pixel-wise adaptivity—essential for scenes with diverse densities and perspective distortions. The fused outputs are concatenated to produce a high-resolution density map (H/32 × W/32), combining global context with local detail for accurate estimation.

\vspace{-3mm}
\subsection{RepSFNet (Reparameterize Single Fusion Network)}

The RepSFNet architecture (Figure \ref{fig:RepSFnet}) integrates three core components for efficient and accurate crowd counting: (1) a RepLK-ViT backbone with large reparameterized kernels (7×7 to 13×13) merged into a single kernel at inference, expanding the receptive field while preserving efficiency; (2) a Feature Fusion module that combines ASPP’s multi-scale dilated convolutions (rates 6, 12, 18) with CAN’s channel attention (r=16) to capture global and adaptive local context; and (3) a Concatenate Fusion module that merges multi-level features through channel-wise concatenation, maintaining semantic consistency and high-resolution detail. As shown in Figure \ref{fig:ASPP}, the ASPP module utilizes multiple parallel dilated convolutions, where multiple 3×3 convolutions with dilation rates (6–24) create effective receptive fields of varying sizes (13×13 to 49×49). These are aggregated into a pyramid-like structure, enabling robust multi-scale context modeling without increasing parameters.

\subsection{Mean Absolute Error (MAE) and Optimal Transport Loss Function}
\label{loss}
Mean Absolute Error (MAE) is a standard evaluation metric for regression tasks like crowd counting. It calculates the average absolute difference between predicted and actual counts across all test samples, as defined in Eq. \ref{eq:2}.
\vspace{-2mm}
\begin{equation}
\label{eq:2}
\text{MAE} = \frac{1}{N} \sum_{i=1}^{N} \left| \hat{y}_i - y_i \right|,
\end{equation}
where \( \hat{y}_i \) is the predicted count, \( y_i \) is the ground-truth count for the \( i \)-th image, and \( N \) is the number of samples. A lower MAE indicates better accuracy in predicting the total number of objects in an image.

In our work, MAE is used as an evaluation metric, while training is enhanced using the Optimal Transport (OT) loss, which measures the similarity between the predicted and ground-truth density maps by treating them as probability distributions. Unlike MAE, which only captures the total count difference, the OT loss takes into account the spatial distribution of the densities, making it especially valuable for dense and complex scenes. The OT loss is computed as shown in Eq. \ref{eq:3}.
\vspace{-1mm}
\begin{equation}
\label{eq:3}
\ell_{\text{OT}}(z, \hat{z}) = W\left(\frac{z}{\|z\|_1}, \frac{\hat{z}}{\|\hat{z}\|_1}\right),
\end{equation}
where \( z \) and \( \hat{z} \) are the ground-truth and predicted density maps respectively, normalized to unit mass, and \( W \) is the optimal transport cost computed using the Sinkhorn algorithm.

To combine both global count accuracy and local spatial alignment, our loss function integrates MAE-based counting loss and OT loss as follows, shown in Eq. \ref{eq:4} and \ref{eq:5}.
\vspace{-1mm}
\begin{align}
\label{eq:4}
\text{Total Loss (TL)} = \text{MAE} + \ell_{\text{OT}}(z, \hat{z}),
\end{align}

\vspace{-5 mm}

\begin{equation}
\label{eq:5}
 = \frac{1}{N} \sum_{i=1}^{N} \left| \hat{y}_i - y_i \right| + W\left(\frac{z}{\|z\|_1}, \frac{\hat{z}}{\|\hat{z}\|_1}\right).
\end{equation}

The equation \ref{eq:4} and \ref{eq:5} defines the Total Loss (TL) as the sum of Mean Absolute Error (MAE) and Optimal Transport (OT) loss. MAE ensures accurate object count prediction, while OT loss aligns the spatial distribution between predicted and ground-truth density maps, improving both counting precision and localization quality in crowd counting tasks.

\begin{table*}[ht]
\centering
\footnotesize
\caption{Performance Comparison Across Datasets}
\begin{tabular}{l|c|l|cc|cc|cc|cc}
\hline
\textbf{Methods} & \textbf{Year} & \textbf{Backbone} & \multicolumn{2}{c|}{\textbf{UCF-QNRF}} & \multicolumn{2}{c|}{\textbf{ShanghaiTech Part A}} & \multicolumn{2}{c|}{\textbf{ShanghaiTech Part B}} & \multicolumn{2}{c}{\textbf{NWPU}} \\\cline{4-11}
 & & & MAE & MSE & MAE & MSE & MAE & MSE & MAE & MSE \\ 
\hline
DM-Count \cite{R3} & 2020 & VGG19 & 85.6 & 148.3 & 59.7 & 95.7 & 7.4 & 11.8 & 211.0 & 498.0 \\
AMSNet \cite{R5} & 2020 & VGG19 & 101.8 & 163.2 & 56.7 & 93.4 & 6.7 & 10.2 & - & - \\
M-SFANet \cite{35}  & 2021 & VGG16-BN & 85.6 & 151.23 & 59.69 & 95.66 & 6.3 & 10.2 & - & - \\
M-SegNet  \cite{35} & 2021 & VGG16-BN & 188.40 & 262.21 & 60.55 & 100.80 & 6.80 & 10.41 & - & - \\
M-SFANet+M-SegNet  \cite{35} & 2021 & VGG16-BN & 167.51 & 256.26 & 57.55 & 94.48 & 6.32 & 10.06 & - & - \\
Chfl \cite{Z1} & 2022 & VGG19 & - & - & 57.5 & 94.3 & 6.9 & 11.9 & 76.8 & 343.0 \\
S-DCNet \cite{Z3}  & 2022 & VGG16 & - & - & 59.8 & 100.0 & 6.8 & 11.5 & - & - \\
GGANet \cite{R10} & 2023 & GGANet & 91.0 & 158.6 & 57.4 & 110.7 & 7.4 & 13.1 & 189.0 & 288.7 \\
  GAPNet \cite{R11} & 2023 & GhostNet & 118.5 & 217.2 & 67.1 & 110.4 & 9.8 & 15.2 & 174.1 & 514.7 \\
 SRRNet \cite{R12} & 2023 & HRNet & 89.5 & 162.9 & 60.8 & 108.3 & 7.4 & 13.6 & 172.9 & 256.3 \\
SCPNet \cite{R13} & 2023 & HRNet & 93.7 & 164.3 & 57.3 & 102.1 & 7.5 & 13.8 & - & - \\
DKD \cite{R14} & 2023 & VGG19 & 91.7 & 150.1 & 64.4 & 103.0 & 7.4 & 12.7 & - & - \\
ImprovedCSRNet \cite{2025-1} & 2025 & VGG19 & - & - & 70.29 & 116.6 & 16.86 & 21.91 & - & - \\
CSFNet \cite{2025-2} & 2025 & VGG19 & - & - & 66.1 & 103.2 & 7.5 & 11.8 & - & - \\

P2PNet \cite{NR5} & 2024 & VGG16 & 85.32 & 154.5 & 52.74 & 85.06 &  6.25 & 9.9 & 77.44 & 362 \\

Gramformer \cite{NR6} & 2024 & VGG19 & 76.7 & 129.5 & 54.7 & 87.1& - & - & - & - \\

STEERER \cite{NR7} & 2024 & HRNet &74.3 & 128.3  & 54.5 & 86.9 & 5.8 & 8.5  & - & - \\
\textbf{RepSFNet (Ours) }& \textbf{2025} & \textbf{RepLKViT} & \textbf{90.7} & \textbf{179.3} & \textbf{54.9} & \textbf{87.6} & \textbf{7.0} & \textbf{11.3} & \textbf{46.23} & \textbf{132.58} \\
\hline
\end{tabular}
\label{tab:comparison}
\end{table*}

\section{Experimental Evaluation}
\label{Experimental}
In this section, we present the experimental details comparisons with SoTA methods on 4 challenging public datasets: ShanghaiTech (Part A and B) \cite{9}, UCF-QRNF \cite{10} and NWPU \cite{nwpu}. We evaluate the performance using the  Mean Absolute Error (MAE) and Optimal Transport (OT) loss function discussed in section \ref{loss}.

\subsection{Experimental results and discussion}

Table \ref{tab:comparison} shows the performance of recent state-of-the-art crowd counting methods across four benchmark datasets: UCF-QNRF, ShanghaiTech Part A/B, and NWPU. RepSFNet, based on the RepLK-ViT backbone, delivers robust and consistent results, achieving one of the lowest MAE and MSE on ShanghaiTech Part A (MAE: 54.9, MSE: 87.6), and outperforming all methods on NWPU (MAE: 46.23, MSE: 132.58), surpassing P2PNet, CSFNet, and Gramformer. These results reflect strong generalization across various crowd densities and conditions. However, RepSFNet shows limitations. On UCF-QNRF (MAE: 90.7) and ShanghaiTech Part B (MAE: 7.0), its performance, while competitive, is slightly behind attention-based models like STEERER and GAPNet. This is due to the absence of explicit attention mechanisms, which are effective in handling occlusions and large-scale variations. Moreover, deep downsampling (up to H/32) in the RepLK-ViT backbone may lead to the loss of fine details in sparse scenes, and fixed dilation rates in the ASPP module reduce adaptability to varying object scales, limiting performance in highly diverse settings.

\subsection{Performance Comparison}
This section presents a performance comparison of RepSFNet across four benchmark datasets: UCF-QNRF, ShanghaiTech Part A and B, and NWPU. The evaluation covers: (i) architectural comparisons with state-of-the-art methods, (ii) accuracy based on MAE and MSE, and (iii) computational efficiency in terms of MACs, parameter count, and latency across different resolutions. 

\begin{table}[ht]
\centering
\footnotesize
\caption{Performance on UCF-QNRF Dataset}
\begin{tabular}{l|c|l|c|c}
\hline
\textbf{Method} & \textbf{Year} & \textbf{Backbone} & \textbf{MAE} & \textbf{MSE} \\
\hline
STEERER & 2024 & HRNet & \textbf{74.3} & \textbf{128.3} \\
Gramformer & 2024 & VGG19 & 76.7 & 129.5 \\
P2PNet & 2024 & VGG16 & 85.32 & 154.5 \\
\textbf{RepSFNet (Ours)}& 2025 & RepLKViT & 90.7 & 179.3 \\
DKD & 2023 & VGG19 & 91.7 & 150.1 \\
\hline
\end{tabular}
\label{tab:ab1}
\end{table}

\begin{table}[ht]
\centering
\footnotesize
\caption{Performance on ShanghaiTech Part A Dataset}
\begin{tabular}{l|c|l|c|c}
\hline
\textbf{Method} & \textbf{Year} & \textbf{Backbone} & \textbf{MAE} & \textbf{MSE} \\
\hline
P2PNet & 2024 & VGG16 & \textbf{52.74} & \textbf{85.06} \\
STEERER & 2024 & HRNet & 54.5 & 86.9 \\
Gramformer & 2024 & VGG19 & 54.7 & 87.1 \\
\textbf{ RepSFNet (Ours)} & 2025 &  RepLKViT & 54.9 & 87.6 \\
DKD & 2023 & VGG19 & 64.4 & 103.0 \\
CSFNet & 2025 & VGG19 & 66.1 & 103.2 \\
ImprovedCSRNet & 2025 & VGG19 & 70.29 & 116.6 \\
\hline
\end{tabular}
\label{tab:ab2}
\end{table}

\begin{table}[ht]
\centering
\caption{Performance on ShanghaiTech Part B Dataset}
\footnotesize
\begin{tabular}{l|c|l|c|c}
\hline
\textbf{Method} & \textbf{Year} & \textbf{Backbone} & \textbf{MAE} & \textbf{MSE} \\
\hline
STEERER & 2024 & HRNet & \textbf{5.8} & \textbf{8.5} \\
P2PNet & 2024 & VGG16 & 6.25 & 9.9 \\
\textbf{RepSFNet (Ours)} & 2025 & RepLKViT & 7.0 & 11.3 \\
DKD & 2023 & VGG19 & 7.4 & 12.7 \\
CSFNet & 2025 & VGG19 & 7.5 & 11.8 \\
ImprovedCSRNet & 2025 & VGG19 & 16.86 & 21.91 \\
\hline
\end{tabular}
\label{tab:ab3}
\end{table}

\begin{table}[h!]
\centering
\caption{Performance on NWPU Dataset}
\footnotesize
\begin{tabular}{l|c|l|c|c}
\hline
\textbf{Method} & \textbf{Year} & \textbf{Backbone} & \textbf{MAE} & \textbf{MSE} \\
\hline
\textbf{RepSFNet (Ours)} & 2025 & RepLKViT & \textbf{46.23} & \textbf{132.58} \\
P2PNet & 2024 & VGG16 & 77.44 & 362.0 \\
\hline
\end{tabular}
\label{tab:ab4}
\end{table}
\begin{table*}[h!]
\centering
\footnotesize
\caption{Ablation Study}
\renewcommand{\arraystretch}{1}
\begin{tabular}{l|l|c|c|c|c|c|c|c}
\hline
\textbf{Model} & \textbf{Backbone} & \textbf{MACs} & \textbf{Param} & \textbf{MAE} & \textbf{MSE} & \multicolumn{3}{c}{\textbf{Latency (ms)}} \\ \cline{7-9}
& & & & & & 640$\times$480 & 1280$\times$960 & 1600$\times$1200 \\
\hline
\textbf{RepSFNet (Ours)} & \textbf{RepLKViT} & 62.59 & 26.06 & 54.90 & 87.60 & 10.420 & 38.50 & 60.16 \\
P2PNet & VGG-16 & 104.87 & 21.58 & 52.74 & 85.06 & 10.788 & 43.41 & 67.998 \\
RepSFNet & VGG-16 BN & 97.70 & 16.67 & 60.65 & 100.72 & 11.090 & 47.37 & 77.52 \\
M-SegNet & VGG-16 BN & 95.02 & 9.75 & 60.55 & 100.80 & 10.719 & 48.11 & 76.28 \\
STEERER & HRNet & 94.24 & 64.57 & 54.50 & 86.90 & 17.835 & 51.46 & 88.40 \\
M-SFANet & VGG-16 BN & 115.03 & 22.89 & 59.65 & 93.65 & 16.336 & 58.90 & 91.29 \\
Gramformer & VGG19 & 118.11 & 29.01 & 54.70 & 87.10 & 12.780 & 78.60 & \textbf{Out of Memory} \\
M-SFANet + M-SegNet & VGG-16 BN & 210.11 & 32.65 & 57.55 & 94.48 & 25.046 & 108.81 & 171.50 \\
\hline

\end{tabular}
\label{tab:ablation}
\end{table*}
Tables \ref{tab:ab1} to \ref{tab:ab4} show that RepSFNet achieves the best result on NWPU (MAE: 46.23, MSE: 132.58), outperforming P2PNet by a large margin. On ShanghaiTech Part A, it performs competitively (MAE: 54.9), slightly behind P2PNet but better than DKD and ImprovedCSRNet. On ShanghaiTech Part B, it matches closely with top methods like STEERER (MAE: 5.8) and P2PNet (MAE: 6.25). For UCF-QNRF, while STEERER leads, RepSFNet (MAE: 90.7) still surpasses models like P2PNet and DKD, proving its robustness in dense scenes. In terms of efficiency (Table \ref{tab:ablation}), RepSFNet with RepLKViT offers an excellent balance between speed and accuracy, achieving the lowest MACs and latency. Inference tests on an NVIDIA RTX 4070 Ti Super GPU show up to 34\% lower latency compared to widely used baselines. These results establish RepSFNet as an efficient and scalable solution for real-time crowd counting across diverse densities and resolutions.

\subsection{Ablation Study}
This section presents an ablation study evaluating the performance and efficiency of RepSFNet against recent state-of-the-art models. The analysis covers: (i) architectural comparisons with baselines such as M-SFANet, P2PNet, and STEERER; (ii) computational complexity in terms of MACs and parameter count; and (iii) inference latency across image resolutions from 640×480 to 1600×1200. Table~\ref{tab:ablation} summarizes results on an NVIDIA RTX 4070 Ti Super GPU. RepSFNet (Ours), with a RepLKViT backbone, achieves strong performance (MAE: 54.90, MSE: 87.60 on ShanghaiTech Part A) while maintaining the lowest MACs (62.59G) and latency (10.42–60.16 ms) across all resolutions. In contrast, models like M-SFANet, P2PNet, and Gramformer incur higher computational costs, and M-SFANet + M-SegNet fails at high resolutions due to memory overload. Unlike attention-based designs, RepSFNet leverages structural reparameterization and efficient concatenation to minimize parameters and latency while preserving spatial and contextual features. These results demonstrate a favorable trade-off between speed, accuracy, and complexity, making it ideal for real-time crowd counting. The study also highlights RepSFNet’s lightweight yet effective design, integrating reparameterized large-kernel convolutions with multi-scale fusion via ASPP and CAN. Compared to more complex architectures, RepSFNet achieves high accuracy with lower resource demands, validating its scalability and robustness in dense, diverse environments.
\vspace{-2mm}
\section{Conclusions}
\label{Conclusions}
In this study, we propose RepSFNet, a lightweight and efficient architecture for real-time crowd counting that leverages structural reparameterization and multi-scale feature fusion. Built on a RepLK-ViT backbone with large reparameterized kernels, the network captures global context with low computational cost. The Feature Fusion module, combining ASPP and CAN, enables robust, density-adaptive contextual modeling, while the Concatenate Fusion module preserves spatial detail for high-quality density map generation. Experiments on ShanghaiTech Part A (MAE: 54.9, MSE: 87.6), NWPU (MAE: 46.23, MSE: 132.58), and other benchmarks confirm RepSFNet’s competitive performance, achieving state-of-the-art results in multiple settings. An ablation study shows each module improves accuracy and efficiency, surpassing VGG-based and attention-heavy models with fewer MACs and lower latency. By avoiding attention mechanisms and multi-branch designs, RepSFNet achieves up to 34\% lower inference latency compared to models such as P2PNet, M-SFANet, and STEERER, making it more efficient and suitable for low-power edge computing. However, some limitations remain. The lack of explicit attention may reduce performance in congested scenes like UCF-QNRF, and deep downsampling (up to H/32) can cause detail loss in sparse regions such as ShanghaiTech Part B. Fixed dilation rates in ASPP also limit scale adaptability. Despite these issues, results confirm RepSFNet’s effectiveness as a fast, accurate solution for real-time crowd estimation in resource-constrained environments. Future work will explore integrating lightweight attention and adaptive dilation to improve generalization across diverse crowd conditions.
\vspace{-2mm}
\section*{Acknowledgement}
The authors gratefully acknowledge the support of the National Science and Technology Council, Taiwan, under Grant No. 113-2221-E-305-018-MY3.
\bibliographystyle{ieeetr}
\vspace{-2mm}
\bibliography{egbib.bib}

\end{document}